\documentclass[letterpaper]{article} 
\usepackage{aaai24}
\usepackage{times}  
\usepackage{helvet}  
\usepackage{courier}  
\usepackage[hyphens]{url}  
\usepackage{graphicx} 

\usepackage{threeparttable}
\usepackage{color}
\usepackage{booktabs}
\usepackage{amsopn}
\usepackage{multirow}
\usepackage{colortbl}
\usepackage{makecell}
\usepackage{tabularx}
\usepackage{bbding}
\usepackage{amsmath}
\usepackage{amssymb}
\usepackage{amsfonts}
\usepackage{bm}
\usepackage{array}
\usepackage{appendix}
\usepackage{adjustbox}

\usepackage{natbib}
\usepackage{caption}
\usepackage{algorithm}
\usepackage{algorithmic}

\usepackage{newfloat}
\usepackage{listings}
\DeclareCaptionStyle{ruled}{labelfont=normalfont,labelsep=colon,strut=off}
\floatstyle{ruled}
\newfloat{listing}{tb}{lst}{}
\floatname{listing}{Listing}

\title{
Object-aware Adaptive-Positivity Learning for Audio-Visual Question Answering 
}

\author{
    Zhangbin Li\textsuperscript{\rm 1},
    Dan Guo\textsuperscript{\rm 1,2,3}\thanks{Corresponding authors},
    Jinxing Zhou\textsuperscript{\rm 1$\ast$},
    Jing Zhang\textsuperscript{\rm 1},
    Meng Wang\textsuperscript{\rm 1,2}
}
\affiliations{
    \textsuperscript{\rm 1}School of Computer Science and Information Engineering, Hefei University of Technology\\
    \textsuperscript{\rm 2}Institute of Artificial Intelligence, Hefei Comprehensive National Science Center   \\ 
    \textsuperscript{\rm 3}Anhui Zhonghuitong Technology Co., Ltd\\
    lizhangbin.mail@gmail.com, guodan@hfut.edu.cn,\\ zhoujxhfut@gmail.com, hfutzhangjing@gmail.com, wangmeng@hfut.edu.cn
}

\usepackage{bibentry}

\begin{document}

\maketitle

\begin{abstract}
This paper focuses on the Audio-Visual Question Answering (AVQA) task that aims to answer questions derived from untrimmed audible videos.
To generate accurate answers, an AVQA model is expected to find the most informative audio-visual clues relevant to the given questions. In this paper, we propose to explicitly consider fine-grained visual objects in video frames (object-level clues) and explore the multi-modal relations (\textit{i.e.}, the object, audio, and question) in terms of feature interaction and model optimization. For the former, we present an end-to-end object-oriented network that adopts a question-conditioned clue discovery module to concentrate audio/visual modalities on respective keywords of the question and designs a modality-conditioned clue collection module to highlight closely associated audio segments or visual objects. For model optimization, we propose an object-aware adaptive-positivity learning strategy that selects the highly semantic-matched multi-modal pair as \textit{positivity}. Specifically, we design two object-aware contrastive loss functions to identify the highly relevant question-object pairs and audio-object pairs, respectively. These selected pairs are constrained to have larger similarity values than the mismatched pairs. The positivity-selecting process is adaptive as the positivity pairs selected in each video frame may be different. These two object-aware objectives help the model understand \textit{which objects are exactly relevant to the question} and \textit{which are making sounds}. Extensive experiments on the MUSIC-AVQA dataset demonstrate the proposed method is effective in finding favorable audio-visual clues and also achieves new state-of-the-art question-answering performance.
The code is available at {https://github.com/zhangbin-ai/APL}.
\end{abstract}

\section{Introduction}

\begin{figure}[!t]
\centering
 \includegraphics[width=0.9\columnwidth]{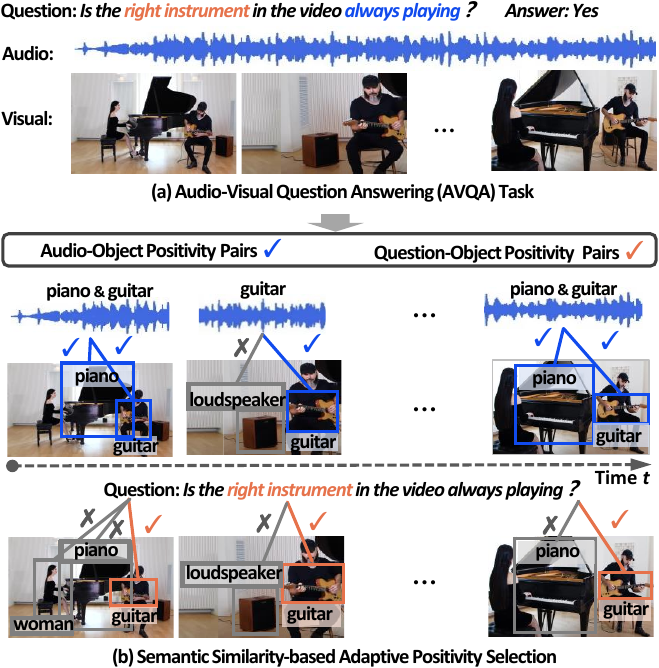}
\caption{
\textbf{Illustration of the AVQA task and our adaptive-positivity selection strategy.}
 (a) The AVQA task requires answering the question with effective audio and visual clues. 
 (b) We explicitly explore the fine-grained visual \textit{objects} and use them as the bridge to adaptively select the highly matched audio-object pairs ({\color{blue}$\checkmark$}) and question-object pairs (\textcolor[RGB]{237,125,49}{$\checkmark$}) by measuring their semantic similarities. In this way, our method is object-aware, which not only knows which objects are making sounds (audible objects) but also identifies which audible objects are really relevant to the question.
 }
 \label{Fig:intro}
\end{figure}

Question-answering occupies a large part of human communication.
To allow machines to learn to respond to questions posed by humans, 
Li \textit{et al.}~\cite{li2022learning} recently proposed the Audio-Visual Question Answering (AVQA) task, which requires answering questions considering both audio and visual information.
Unlike other audio-visual tasks, such as the audio-visual segmentation ~\cite{zhou2022audio,zhou2023audio,mao2023multimodal,li2023transformer,li2023vigt,li2021proposal} and audio-visual event localization~\cite{tian2018audio,zhou2021positive,zhou2022cpsp,zhou2023improving} independently explore the \textit{spatial} and \textit{temporal} relations between audio and visual modalities, the AVQA task requires comprehensive spatial-temporal reasoning of the audio-visual scenarios.
As shown in Fig.~\ref{Fig:intro} (a), to answer the question ``\textit{Is the right instrument in the video always playing?}'', an AVQA model should identify the object guitar at the right, \textit{i.e.}, its visual appearance and \textit{spatial} location, and listen to the audio of all the \textit{temporal} segments.

Previous AVQA works~\cite{li2022learning, lao2023coca} usually adopt a two-stage strategy that first trains an audio-visual model to locate the sounding visual regions (from a spatial perspective) and then uses the question to highlight key temporal audio-visual segments (from a temporal perspective). 
Although this strategy is somewhat helpful for spatial-temporal reasoning in audio-visual scenarios, the sounding visual regions highlighted by the independent first stage may not relate to the question and the second stage can focus on the visual information related to the question only at the segment level.
In this work, we simultaneously consider object-based spatial and temporal learning. This is inspired by following observations: 1) objects are fundamental units in human cognition~\cite{spelke2007core},
2) objects can be a bridge since a question usually focuses on specific objects, and the audio is also emitted by specific objects, and 3) it is easy to spatially differentiate objects in temporal visual segments and the objects convey clear semantics.
Therefore, we propose to explicitly exploit the basic elements -- objects, seeking fine-grained clues from the object level for the AVQA task.
Although some video QA works~\cite{ijcai2021p88, xiao2022video} also use visual objects, 
the audible objects are still in fancy and not explored in the studied AVQA task.
In other words, we need to identify \textit{which objects are corresponding to the audio} and \textit{which are really relevant to the question} in each video frame.

To achieve this goal, an AVQA model needs to first understand the question accurately and know what information to look for in audio and visual modalities.
Next, the model is expected to find informative clues from audio and visual modalities to answer the question.
This two-step decision-making process mimics how humans answer a question.
As the example shown in Fig.~\ref{Fig:intro} (a), 
given the question, we highlight the keywords ``\textit{right instrument}'' and start to notice the corresponding \textit{guitar} at the right position when watching the visual frames. Meanwhile, we listen to the entire audio to judge if the guitar is ``\textit{always playing}''.
After going through the entire video, we pick out the audio-visual clues and find that they satisfy the question, giving the answer ``\textit{Yes}''.
Motivated by this, we design a Question-conditioned Clue Discovery (QCD) module and a Modality-conditioned Clue Collection (MCC) module.
QCD is used to seize the keywords in the question related to audio and visual modalities respectively, and MCC is used to enhance the key audio-visual clues, \textit{e.g.}, the relevant objects and audio segments.
The QCD and MCC modules are equipped with the Transformer~\cite{vaswani2017attention} architecture to highlight key clues via cross-modal attention.
These two modules form our backbone network for encoding the multi-modal relations among the audio, objects, and question.

More importantly, we propose an object-aware adaptive-positivity learning strategy to guide the network to enhance cross-modal features.
This is inspired by the observation that the video frames contain multiple objects, whereas some of them may not be relevant to the question (especially the ones not mentioned in the question) or the audio (silent objects).
As shown in Fig.~\ref{Fig:intro} (a), for the second video segment, the question cares about the ``\textit{right instrument}'' and only the \textit{guitar} is making sound, thereby the detected ``\textit{loudspeaker}'' is a distraction since it does not match the question and audio.
We define the semantic-matched question-object pair or audio-object pair as \textit{positivity}.
These positivity pairs should be highlighted and distinguished from the mismatched pairs for a better understanding of the audio-visual scenarios.
We illustrate the positivity selection process in Fig.~\ref{Fig:intro} (b).
For each video segment, we calculate the similarity values between the features of the question-object pairs and the audio-object pairs.
The question-object positivity pairs (\textcolor[RGB]{237,125,49}{$\checkmark$}) and audio-object positivity pairs ({\color{blue}$\checkmark$}) can be adaptively distinguished if their similarity values surpass a pre-set threshold, while the remaining pairs are merged into a negativity set.
For model optimization, our positivity learning strategy adopts two object-aware contrastive loss functions to respectively constrain the question-object pairs and audio-object pairs in the positivity set to be close, while the unmatched ones in the negativity set are far away.
Note that this contrastive learning scheme can be applied in each video segment, the positivity pairs can be adaptively selected from both spatial and temporal perspectives. 
As the example shown in Fig.~\ref{Fig:intro} (b), multiple objects are treated as positivities if they are relevant to the question/audio (from a spatial perspective at current frame) and simultaneously exist in the changeable positivity object sets along the timeline (from a temporal perspective in the video).
As a result, the model is able to recognize the objects mostly related to the question and audio in each segment, achieving better audio-visual scene understanding.

We refer to the full methodology as Adaptive-Positivity Learning (APL) and evaluate it on the widely-used MUSIC-AVQA ~\cite{li2022learning} dataset and the experimental results demonstrate that our method is able to identify informative audio-visual clues for better question answering.
In summary, we have the following contributions:
\begin{itemize}
    \item We are the first to explore the fine-grained audible object clues in the AVQA task and present an object-oriented network that effectively encodes the multi-modal relations among the objects, audio, and question. 
    \item We propose an object-aware adaptive-positivity learning strategy including two contrastive loss objectives that adaptively highlight the relevant question-object and audio-object pairs for better model optimization.
    \item
    Extensive experimental results show that our method is effective and achieves new state-of-the-art performance on the MUSIC-AVQA dataset.
\end{itemize}

\begin{figure*}[!t]
     \centering 
     \includegraphics[width=1\textwidth]{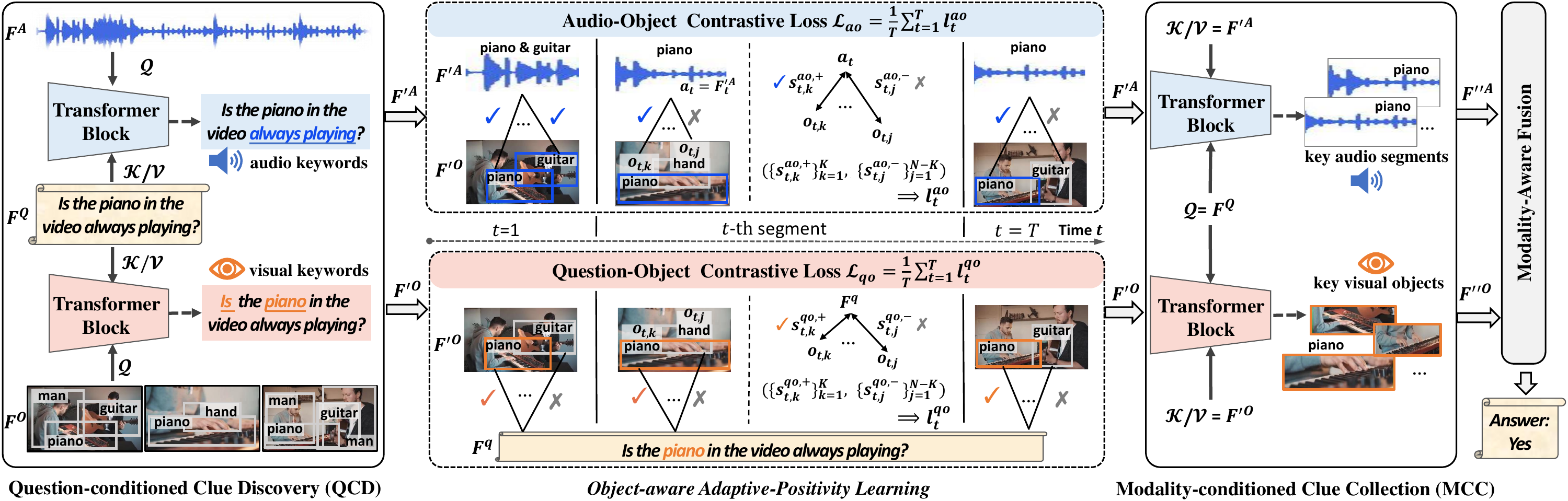}
     
     \caption{\textbf{Method Overview.}
     Given the extracted audio/object/question features ($\bm{F}^{A/O/Q}$), we design two modules to build the backbone network for multimodal interaction.
     The \emph{Question-conditioned Clue Discovery ({{QCD}})} module individually employs different modalities, \textit{i.e.}, audio and visual (in-frame objects), to highlight the responsive keywords of the question (Left). Then, the \emph{Modality-conditioned Clue Collection ({{MCC}}}) module collects relevant audio and visual clues for answering the question (Right). Between these two modules, we design a novel \textit{Object-aware Adaptive-Positivity Learning} strategy that contains two contrastive loss functions to adaptively recognize the mostly relevant question-object and audio-object positivity pairs.
     This object-aware design facilitates informative cross-modal feature encoding and better model optimization.
     The learned cross-modal features are finally input to a \textit{Modality-aware Fusion} module to balance the audio-visual clues for answer prediction.
     }
     \label{fig:framework}
\end{figure*}

\section{Related Work}
\textbf{Question Answering in Different Modalities.}
Intelligent questioning and answering (QA) tasks originated from text-based QA ~\cite{mishra2016survey,song2022memorial} and are increasingly incorporating more and more modality data such as text, images, video, and audio, resulting in Visual QA (VQA)~\cite{ben2017mutan, liu2021dual,zhou2023exploring}, Video QA\cite{lei2018tvqa, shen2023fine}, Audio QA~\cite{chuang2019speechbert, li2023multi}, and the recently emerged Audio-Visual QA (AVQA)~\cite{yun2021pano, li2022learning, lao2023coca} tasks. 
Among these tasks, the AVQA task in this study is more challenging, as it requires synthesizing audiovisual information to answer the questions, involving more sophisticated cross-modal semantic understanding and spatio-temporal reasoning.

\noindent\textbf{Audio-Visual Question Answering.}
In the AVQA task, existing works focus on the perception learning of cross-modalities. AVSD~\cite{schwartz2019simple} incorporates a multi-modal cross-modal attention mechanism and an LSTM module for answer generation. Pano-AVQA~\cite{yun2021pano} studies the audiovisual question-answering in panoramic video scenarios. They propose a transformer-based method that adopts multiple transformer encoders to integrate the cross-modal relations in audio, visual, and language (question).
The questions in Pano-AVQA are generally simple and the videos are short (5 seconds). Later, Li \textit{et al.} \cite{li2022learning} build a new MUSIC-AVQA dataset in long music scenarios and propose a two-stage method. A spatial grounding module is first used to highlight the visual regions corresponding to the audio. Then, the updated visual and audio features are sent to a temporal grounding module that uses the question to generate modality-related attention weights.
Lao \textit{et al.}~\cite{lao2023coca} adopt the same framework but utilize the casual analysis to alleviate the shortcut bias problem. 
Unlike these works, our method explores fine-grained visual objects and activates the audio-matched and question-matched objects to learn multimodal relations. This helps the model to gather more precise audio-visual clues related to the question thus giving better answers.

\section{Methodology}

As our model is an object-aware solution, we focus on the exploitation of the object semantics.
We first introduce the backbone network that incorporates the object, audio, and question for multimodal feature interaction.
Then, we elaborate on the proposed object-aware adaptive-positivity learning strategy that regularizes the question-object and audio-object semantic alignments for better feature encoding and model optimization.

\subsection{Backbone Network}
\label{sec:pipeline}
The main modules of the proposed network are shown in Fig.~\ref{fig:framework}.
Given an audible video, we first down-sample it to $T$ segments and then extract its audio feature and object feature, denoted as $\bm{F}^A \in \mathbb{R}^{T \times d}$ and $\bm{F}^O \in \mathbb{R}^{T \times N \times d}$, respectively. $N$ is the total number of detected objects in one video frame, and $d$ is the feature dimension.
For the raised question, we extract the word-level feature $\bm{F}^Q \in \mathbb{R}^{L \times d}$ and the sentence-level feature $\bm{F}^q \in \mathbb{R}^{1 \times d}$ using an LSTM, where $L$ is the word length of the question.
Then, the extracted multimodal features are processed by our multi-modal interaction scheme that contains two modules to effectively discover informative audio-visual clues by encoding the cross-modal relations with the question.
Specifically, the \textit{Question-conditioned Clue Discovery (QCD)} module encourages the audio/visual modality to focus on relevant keywords in the question.
Then, the \textit{Modality-conditioned Clue Collection (MCC)} module enhances the features from informative audio segments and visual objects.
After that, all the visual, audio, and textual features are incorporated to predict the answer in the \textit{Modality-Aware Fusion} module. 
We elaborate on the details of each module below.

\noindent\textbf{{Question-conditioned Clue Discovery (QCD).}}
To exactly answer a question, each modality should emphasize more on the relevant keywords in the question.
As shown in Fig.~\ref{fig:framework}, there is a question ``\textit{Is the piano in the video always playing?}''. The audio modality should focus on the words ``\textit{always playing}'' in the question, and the visual modality should pay strong attention to the word ``\textit{piano}''. 
This inspires us to design a question-conditioned clue discovery module.
Specifically, we use the Transformer encoder ({TFM})~\cite{vaswani2017attention}
to encode the cross-modal relations of the audio or visual objects to the question. The TFM mainly consists of a multi-head attention ({MHA}) and a feed-forward network ({FF}). Overall, the TFM can be defined as: 
\begin{equation}
\begin{aligned}
{\rm TFM} (\bm{\mathcal{Q}}, \bm{\mathcal{K}}, \bm{\mathcal{V}})
= {\rm {FF}}({\rm {MHA}}(\bm{\mathcal{Q}},\bm{\mathcal{K}},\bm{\mathcal{V}})), \\
\end{aligned}
\label{eq:TFM}
\end{equation}
where $\bm{\mathcal{Q}}$, $\bm{\mathcal{K}}$, and $\bm{\mathcal{V}}$ denotes the \textit{query}, \textit{key}, and \textit{value}, respectively. 
Then, the QCD process is summarized as:
\begin{equation}
\begin{aligned}
\bm{F}^{'m} = {\rm TFM} (\bm{F}^{m},\bm{F}^{Q},\bm{F}^{Q}), \\
\end{aligned}
\label{eq:QCD}
\end{equation}
where $m \in \{O, A\}$ denotes the object feature and audio feature. Hence, we obtain the enhanced visual object feature $\bm{F}^{'O} \in \mathbb{R}^{T \cdot N \times d}$ and the audio feature $\bm{F}^{'A} \in \mathbb{R}^{T \times d}$.

\noindent\textbf{{Modality-conditioned Clue Collection (MCC).}}
MCC aims to further collect all the useful audio-visual clues related to the question from each modality for effective answering.
The implementation of MCC is similar to QCD but the question is used as the \textit{query} in the multi-head attention.
The calculation process is formulated as:
\begin{equation}
\begin{aligned}
\bm{F}^{''m} = {\rm TFM} (\bm{F}^{Q},\bm{F}^{'m},\bm{F}^{'m}), \\
\end{aligned}
\label{eq:MCC}
\end{equation}
where $m \in \{O, A\}$ denotes the object or audio modality. Until now, we update the question-related clues from each modality and obtain the corresponding cross-modal features, denoted as $\bm{F}^{''O} \in \mathbb{R}^{L \times d}$  and $\bm{F}^{''A} \in \mathbb{R}^{L \times d}$.

\noindent\textbf{Modality-Aware Fusion.}
In general, both audio and visual modalities provide useful information for predicting answers. However, some types of questions may be more closely related to a particular modality.
The AVQA model should be modality-aware when gathering audio-visual clues.
We achieve this goal by generating a modality-aware weight vector $\bm{\beta} \in \mathbb{R}^{1 \times 2}$ from the sentence-level question feature $\bm{F}^q$ using a linear layer followed by a \textit{softmax} activation. The object-related feature $\bm{F}^{''O}$ and audio-related feature $\bm{F}^{''A}$ are then transformed and weighted by $\bm{\beta}$ to obtain the final fused feature $\bm{f}_{out} \in \mathbb{R}^{1 \times d}$.\\
\noindent{\textit{Answer prediction.}} The AVQA~\cite{li2022learning} is performed in an open-ended manner, which aims to select the correct answer from a pre-defined answer vocabulary.
To this end, we send the final output vector $\bm{f}_{out}$ to a linear layer activated by \textit{softmax} function for answer prediction.
This generates an answer probability vector $\bm{p} \in  \mathbb{R}^C$, where $C$ is the number of candidate answers.
During inference, the predicted answer is decided by $\hat{c} = arg~\text{max}_c(\bm{p})$.

\subsection{Object-aware Adaptive-Positivity Learning}\label{sec:contrastive}
Existing methods~\cite{li2022learning, lao2023coca} usually adopt the cross-entropy loss between the prediction and the ground truth answer to supervise model training.
In this work, we emphasize that an AVQA model should understand the \textit{semantic positivity} contained in the multi-modalities for better question answering.
On one hand, {those objects relevant to the question should be highlighted}.
The video frames usually contain many objects whereas some of them may be distractions as they do not match the question. 
On the other hand, those objects corresponding to the audio signal may also be considered. The audio-object semantic matched pairs would also help the model understand the audio-visual scene, \textit{i.e.}, {what the sounding objects are}. We treat these semantically matched question-object and audio-object pairs as \textbf{positivity pairs}.
Then, we design two multimodal positivity learning strategies, one for question-object pair matching and the other for audio-object pair matching. 

Take the question-object positivity learning as an example, for the $t$-th segment, the sentence-level question feature and object feature are $\bm{q} =\bm{F}^{q} \in \mathbb{R}^{1 \times d}$ and $\bm{o}_{t}=\{\bm{o}_{t, i}\}_{i=1}^{N}=\bm{F}_t^{'O} \in \mathbb{R}^{N \times d}$, respectively.
The relevance of the question-object pairs can be measured by the 
cosine similarity $\bm{s}^{{qo}}_t$,
\begin{equation}
    \bm{s}^{{qo}}_t = softmax(\frac{\bm{q}}{ \|\bm{q}\|} \otimes (\frac{\bm{o_t}}{\| \bm{o}_t \|})^{\top}),
\end{equation}
where $\otimes$ is the matrix multiplication, $t=1,2,..., T$,
$\bm{s}_t^{{qo}} \in \mathbb{R}^{1 \times N}$ reflects the similarity of the question with all $N$ detected objects at the $t$-th segment.
A high similarity score of $\bm{s}^{{qo}}_t$, \textit{e.g.}, ${s}^{{qo}}_{t, i}$ indicates that the $i$-th object is more likely to be relevant to the question.

We then use a simple thresholding strategy to select the highly relevant question-object pairs, forming the positivity set, denoted as $\mathcal{P}^{qo}_t$. While the remaining pairs are more likely to be irrelevant, forming the negativity set, denoted as $\mathcal{N}^{qo}_t$. This selection process can be summarized as,
\begin{equation}
\left\{\begin{matrix}
\begin{aligned}
\mathcal{P}^{qo}_t &= \{\bm{s}^{qo}_{t} | \bm{s}^{qo}_{t} > \varphi \},\\
\mathcal{N}^{qo}_t &= \{\bm{s}^{qo}_{t} | \bm{s}^{qo}_{t} \leq \varphi\},
    \end{aligned}
    \end{matrix}\right.
\label{Eq:adptive_param}
\end{equation}
{where $\varphi$ is the threshold, $ \{\cdot \}$ denotes the selected question-object pair set satisfying the condition.
Assume there are $K$ question-object positivity pairs in $\mathcal{P}^{qo}_{t}=\{ {s}_{t,k}^{qo,+} \}_{k=1}^{K}$,  $\mathcal{N}^{qo}_{t}=\{ {s}_{t,j}^{qo,-} \}_{j=1}^{N-K}$ will contain $N-K$ pairs.
$N$ is the total number of detected objects and is fixed, whereas $K$ is dynamic which means the positivity pair selection process is \textbf{{adaptive}}.
This can be reflected in two aspects:
1) At the beginning of model training, a large portion of detected objects even irrelevant to the question may be regarded as positivity pairs. As the training proceeds, the model gradually acquires the ability to preserve mainly the connections between the most relevant question-object pairs. 
For each video frame, the model is able to adaptively recognize key objects at different \textit{spatial} locations.
2) The positivity pair selection can be applied in all the $T$ video segments. For each segment, it encourages the model to identify the question-related objects in the current timestamp.
This is also an adaptive process along the \textit{temporal} dimension.
As the example shown in Fig.~\ref{fig:framework}, our model adaptively selects the ``\textit{piano}'' for all the video frames, while ignoring those objects that are irrelevant to the question, such as ``\textit{hand}'' and ``\textit{guitar}''.

\begin{table*}[t]
  \centering
  \renewcommand{\arraystretch}{1.3}
  \begin{adjustbox}{width=\textwidth}
  \begin{threeparttable}
    \begin{tabular}{c|cc|c|cc|c|ccccc|c|c}
    \Xhline{1.2pt}
    \multirow{2}{*}{Method}&
    \multicolumn{3}{c|}{Audio Question}&\multicolumn{3}{c|}{Visual Question}&\multicolumn{6}{c|}{Audio-visual Question}&All  \\
    \cline{2-14}
    ~ &Counting &Comparative &Avg. &Counting &Location &Avg. &Existential &Location &Counting &Comparative &Temporal &Avg.
    &Avg. \\
   \hline
   
    AVSD~\cite{schwartz2019simple} &72.47 &62.46 &68.78 &66.00 &74.53 &70.31 &80.77 &64.03 & 57.93 &62.85 &61.07 &65.44 & 67.32 \\
    
    Pano-AVQA~\cite{yun2021pano} &75.71 &65.99 &72.13 &70.51 &{75.76} &73.16 &82.09 &65.38 &61.30 &63.67 &62.04 &66.97 &69.53 \\
    
    STG~\cite{li2022learning} &77.78	&67.17	&73.87	&73.52	&75.27	&74.40 	&82.49	&64.24	&{69.88}	&\underline{64.67}	&\underline{65.82} &69.53	&71.59 \\
        
    COCA~\cite{lao2023coca} & {79.94}	&{67.68}	&{75.42}	&{75.10} 	&75.43	&{75.23}	&\textbf{83.50}	& {66.63}	&69.72	&64.12	&65.57	&{69.96}	&{72.33} \\ \hline
        
    \textbf{APL}$\text{-Faster R-CNN}$ (ours) &\underline{81.81}	&\underline{68.86}	&\underline{77.03}	&\underline{76.44}	&\underline{82.29}  &\underline{79.40}	&{82.29}	&\textbf{67.28}	&\underline{72.33}	&63.49 &65.33	&\underline{70.31}	&\underline{73.91} \\
        
    \textbf{APL}$\text{-DETR}$ (ours) &\textbf{82.40} &\textbf{70.71} &\textbf{78.09}	&\textbf{76.52}	&\textbf{82.74} &\textbf{79.69}	&\underline{82.99}	& \underline{66.68}	&\textbf{73.29}	&\textbf{64.76}	&\textbf{65.95}	&\textbf{70.96}	&\textbf{74.53} \\
    \Xhline{1.2pt}
    \end{tabular}
    \end{threeparttable}
    \end{adjustbox}
\caption{\textbf{Comparison with existing methods on the MUSIC-AVQA dataset.} We evaluate the performance of each model on different question types and use the QA accuracy (\%) as the metric. We implement our APL method with different object detectors, \textit{e.g.}, the Faster R-CNN and DETR. The best and second-best results are \textbf{bolded} and \underline{underlined}, respectively.}
\label{tab:sota}
\end{table*}

With the positivity set $\mathcal{P}^{qo}_t$ and negativity set $\mathcal{N}^{qo}_t$, we then design a question-object positivity contrastive loss $\mathcal{L}_{qo}$ to regularize question-object pairs in $\mathcal{P}^{qo}$ to have large cross-modal similarities, while low values for pairs in $\mathcal{N}^{qo}$.
$\mathcal{L}_{qo}$ can be computed by,
\begin{equation}
    \left\{\begin{matrix}
\begin{aligned}
 \mathcal{L}_{qo} &= \frac{1}{T} \sum_{t=1}^{T} l^{qo}_t, \\
    l_t^{qo} &= - \text{log} \frac{\sum_{k=1}^K \text{exp}({s}^{{qo,+}}_{t,k} / \tau)}{\sum_{k=1}^K \text{exp}({s}^{{qo,+}}_{t,k} / \tau) + \sum_{j=1}^{N-K} \text{exp}({s}_{t,j}^{{qo,-}} / \tau)},
    \end{aligned}
    \end{matrix}\right.
\label{Eq:qo_loss}
\end{equation}
where ${s}^{qo,+}_{t,k} \in \mathcal{P}_t^{qo}$, ${s}_{t,j}^{qo,-} \in \mathcal{N}_t^{qo}$,
$\bm{s}^{qo,+}_{t,k}$ denotes the similarity of the $k$-th question-object pair coming from the $\mathcal{P}^{qo}_t$ set.
$\tau$ is a hyper-parameter balancing the loss computation, and $l_t^{qo}$ is the loss term at the $t$-th video segment.

Similarly, we can compute the audio-object positivity contrastive loss $\mathcal{L}_{ao}$ that encourages the model to perceive the semantic-matched audio-object pairs.  
The final positivity learning loss $\mathcal{L}_{pc}$ is the combination of $\mathcal{L}_{qo}$ and $\mathcal{L}_{ao}$: $\mathcal{L}_{pc} = \mathcal{L}_{qo} + \mathcal{L}_{ao}$.
$\mathcal{L}_{pc}$ is then added to the basic cross-entropy loss $\mathcal{L}_{ce}$ as final objective for model optimization:
\begin{equation}\label{Eq:total_loss}
    \mathcal{L} = \mathcal{L}_{ce} + \lambda \mathcal{L}_{pc},
\end{equation}
where $\lambda$ is a constant weight to balance the two loss items.

\section{Experiment}

\subsection{Experimental Setup}

\textbf{Dataset.}
Experiments are conducted on the widely-used MUSIC-AVQA  dataset \cite{li2022learning}, which contains over 45K question-answer pairs distributed across 9,288 videos spanning more than 150 hours.
The dataset encompasses 9 types of QA pairs related to diverse modalities, \textit{i.e.}, the audio-visual (AV-), separate visual (V-), and separate audio (A-), depending on which modalities are used to discover question-related clues for answer prediction.
The audio-visual QA pairs occupy a large portion of the entire dataset, and there are five audio-visual question types referring to \emph{existential}, \emph{counting}, \emph{location}, \emph{comparative}, and \emph{temporal}.
The dataset is split into the training, validation, and test sets, which comprise 32K, 4K, and 8K QA pairs, respectively.

\noindent \textbf{Evaluation Metric.}
Following the protocol \cite{li2022learning, lao2023coca}, we test the model performance on each question type and use the average answer prediction accuracy \textbf{\textit{Avg.}} as the metric, and it can be obtained by dividing the number of correct instances by the total number of questions of each particular type and then averaging.

\noindent \textbf{Implementation Details.}
We down-sample each audible video into $T =10$ non-overlapping segments,
resulting in $T$ audio segments and $T$ frames per video.
For the audio segments, we use the VGGish~\cite{hershey2017cnn} network pretrained on the AudioSet~\cite{gemmeke2017audio} to extract the 128-D features. About the visual object features, we test two popular off-the-shelf object detectors, \textit{i.e.}, Faster R-CNN \cite{ren2015faster}
and DETR \cite{carion2020end} for object detection and extracting corresponding features. {We use independent linear layers to transform both object and audio features into the same 512 dimension.
The detected object number $N$ per frame is 36 for Faster R-CNN and 100 for DETR.
Accordingly, we set $\varphi$ in Eq.~\ref{Eq:adptive_param} to 0.028 and 0.011, respectively. 
During training, the parameter $\tau$ in Eq.~\ref{Eq:qo_loss} is set to 0.4 and $\lambda$ in Eq.~\ref{Eq:total_loss} is set to 0.3.
The initial learning rate is set to $1.75e$-4 when using Faster R-CNN and $1e$-4 for DETR.}
The learning rate decreases by multiplying 0.1 every 8 epochs with the Adam optimizer.
The batch size is set to 64 and we train the model for 20 epochs.

\subsection{Comparison with State-of-the-arts}

We conduct the proposed APL model with the DETR object detector here. An ablation study of different object detectors will be shown in the next section.
The experimental results are shown in Table~\ref{tab:sota}.
We report the model performance on different question types.
\textbf{First}, our method significantly surpasses the previous SOTA method COCA~\cite{lao2023coca} on audio question answering and visual question answering: the corresponding average accuracy values are improved by 2.67\%
and 4.46\%,
respectively.
Questions under these two settings are simpler and only related to a single modality.
The performance improvements indicate the proposed method has a better ability to perceive single-modal scenes.
\textbf{Second}, our method is also superior in more challenging audio-visual question answering. Performance of the most competitive method COCA is close to the previous method STG~\cite{li2022learning}, while our method achieves 1\% improvements.
{These results reveal that our approach is superior in various QA settings.}
Moreover, we can also notice that it is more difficult for a model simultaneously handle different audio-visual question types as they focus on different audio-visual clues.
Particularly, our method brings obvious improvement for the question type of \textit{`counting'}, achieving 73.29\% accuracy which considerably surpasses the second-best method STG (69.88\%). This result verifies the benefits of our object-aware method design.

\begin{figure*}[t]
\centering
  \includegraphics[width=0.83\textwidth]{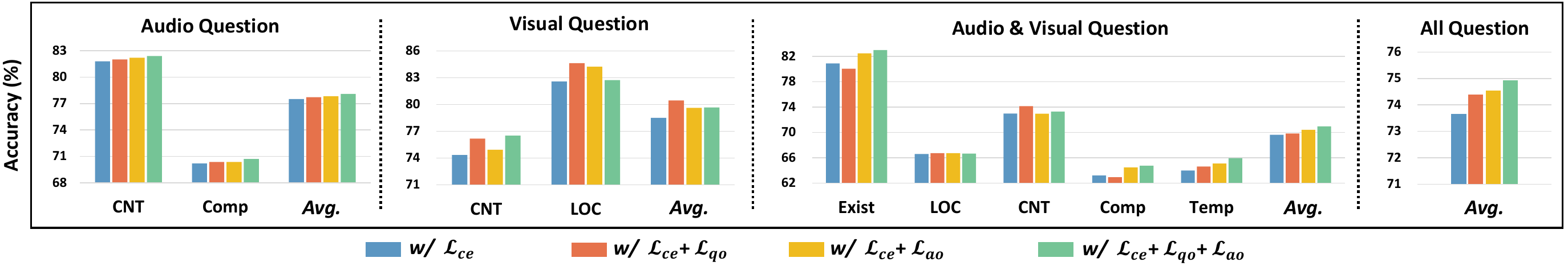}
  \caption{
    \textbf{Ablation study of the adaptive contrastive losses.}
    $\mathcal{L}_{ce}$ computes the cross-entropy loss of the predicted answer and ground truth. $\mathcal{L}_{qo}$ and $\mathcal{L}_{ao}$ are the proposed contrastive losses.
    The `CNT', `Comp', `LOC', `Exist', and `Temp' are the abbreviations for question types `Counting', `Comparative', `Location', `Existential', and `Temporal', respectively.
  }
  \label{Fig:sem_loss}
\end{figure*}

\subsection{Ablation Studies}\label{sec:ablation}

\noindent\textbf{Impacts of Object Detectors.}
To explore the impact of object detectors in our method, here we employ the popular Faster R-CNN and DETR to implement the proposed method, respectively.
As shown in the bottom part of Table~\ref{tab:sota}, our method using the stronger DETR has the best performance, approaching 74.53\% QA accuracy.
Also, it is noteworthy that our method equipped with each object detector is superior to existing methods.
These results demonstrate the generalization and robustness of our method.
In the following experiments, unless otherwise specified, we adopt the DETR as the object detector.

\noindent\textbf{Effects of Main Network Modules.}
Our network uses the question-conditioned clue discovery (QCD) module and modality-conditioned clue collection (MCC) module to encode question-related multi-modal features.
We conduct an ablation study to explore the impacts of these modules.
The experimental results are shown in Table~\ref{tab:modules}.
The first row (\#1) denotes that the QCD and MCC are simply replaced with several fully connected layers to encode features.
The model performance in this extreme case without QCD and MCC modules is very low.
After adopting the QCD or MCC module independently, the model performance improved substantially to $\sim$72\% (\#2 and \#3 in Table).
The model achieves the best performance when using both QCD and MCC modules (\#4).
These results directly show the effectiveness of each proposed module, which is helpful to encode question-related audio-visual features for answer prediction.

\begin{table}[t]
  \centering
  \renewcommand{\arraystretch}{1.2}
  \fontsize{8}{8}\selectfont
  \begin{adjustbox}{width=\columnwidth}
  \begin{threeparttable}
    \begin{tabular}{c | cc|ccc|c}
    \Xhline{1.2pt}
    \multirow{2}{*}{\#} & \multicolumn{2}{c|}{Module} & \multicolumn{4}{c}{Avg. (\%)}  \\
    \cline{2-7}
   ~ &  QCD & MCC  &\multirow{1}{*}{A Question} & \multirow{1}{*}{V Question} & \multirow{1}{*}{A-V Question}  & \multirow{1}{*}{All}\\
    \hline
    1 & - & - &44.01 &18.41 &24.08 &26.09 \\
    
    2 & $\checkmark$ & - &\underline{74.74}  &76.88  &\underline{68.70}  &\underline{71.94} \\
    
    3 & - & $\checkmark$ &73.68 &\underline{77.50} &68.21 &71.64 \\
    
    4 & $\checkmark$ & $\checkmark$ &\textbf{78.09} &\textbf{79.69} &\textbf{70.96} &\textbf{74.53} \\
    \Xhline{1.2pt}
    \end{tabular}
    \end{threeparttable}
    \end{adjustbox}
\caption{\textbf{Ablation study of main network modules.} We report the average accuracy of audio, visual, and audio-visual question answering. `-' denotes that module is not used.}
\label{tab:modules}
\end{table}

\noindent\textbf{Effects of Adaptive Contrastive Losses
$\mathcal{L}_{qo}$ and $\mathcal{L}_{ao}$.}
For model optimization, in addition to the basic cross-entropy loss $\mathcal{L}_{ce}$, we propose two adaptive-positivity contrastive losses, \textit{i.e.}, $\mathcal{L}_{qo}$ and $\mathcal{L}_{ao}$, which encourage the model to be aware of the semantic matched question-object pairs and audio-object pairs, respectively.
We use different loss functions to train the backbone network.
As shown in Fig.~\ref{Fig:sem_loss}, the model generally performs worst for the majority of the question types merely using $\mathcal{L}_{ce}$.
The model performance is improved after combining the proposed  $\mathcal{L}_{qo}$ or $\mathcal{L}_{ao}$ with $\mathcal{L}_{ce}$ for training.
In particular, $\mathcal{L}_{qo}$ lifts the visual QA average accuracy from 78.52\% to 80.47\%, and  $\mathcal{L}_{ao}$ increases the audio-visual QA average accuracy from 69.64\% to 70.43\%.
The QA performance is further boosted when using both contrastive loss items, as shown in the right part of the figure.
These results verify the benefits of the proposed contrastive learning that adaptively identifies the highly relevant question-object and audio-object pairs, facilitating the spatial-temporal reasoning of the audio-visual scenes.

\noindent\textbf{Effects of Threshold $\varphi$ in $\mathcal{L}_{qo}$ and $\mathcal{L}_{ao}$.}
$\varphi$ is the threshold for selecting highly relevant question-object and audio-object positivity pairs for contrastive learning (Eq.~\ref{Eq:adptive_param}).
Here we test several values of $\varphi$ to explore its effect.
As shown in Table~\ref{tab:topk}, the QA accuracy values of most question types change as the $\varphi$ varies.
And the average performance across all question types approaches the highest when $\varphi$ is set to 0.011.
Note that we use the DETR object detector in this experiment and it extracts $N=100$ objects for each video frame. The cross-modal similarity of each question-object (or audio-object) pair will be 0.01 if they share the same attention. 
$\varphi = 0.011$ is slightly larger than this value to ensure that fewer highly matched pairs are selected.
When all the objects are selected as positivities ($\varphi=0$), $\mathcal{L}_{qo}$ (Eq.~\ref{Eq:qo_loss}) and $\mathcal{L}_{ao}$ are equal to 0, the model has sub-optimal performance.

\begin{table}[!t]
  \centering
  \renewcommand{\arraystretch}{1.2}
  \fontsize{8}{8}\selectfont
  \begin{adjustbox}{width=\columnwidth}
  \begin{threeparttable}
    \begin{tabular}{c | c|ccc|c}
    \Xhline{1.2pt}
    \multirow{2}{*}{\#} & \multirow{2}{*}{ Threshold $\varphi$ } & \multicolumn{4}{c}{Avg. (\%)} \\
    \cline{3-6}
    ~ &  ~ &\multirow{1}{*}{A Question} & \multirow{1}{*}{V Question} & \multirow{1}{*}{A-V Question}  & \multirow{1}{*}{All} \\
    \hline
    1 & $\varphi=0 \quad \; \; \;$ &77.52  &78.52  &69.64  &73.39 \\
    2 & $\varphi=0.009$ &77.34  &79.22  &69.74  &73.60 \\
    3 & $\varphi=0.010$ &{77.96}  & \textbf{80.26}  &70.09  &74.18 \\
    4 & $\bm{\varphi}=\bm{{0.011}}$ &\textbf{78.09}  &79.69  &\textbf{70.96}  &\textbf{74.53} \\
    5 & $\varphi=0.012$ &77.72  &79.52  & {70.62}  & {74.24} \\
    6 & $\varphi=0.013$ & 77.65 & {79.85} & 70.04 &73.98 \\
    \Xhline{1.2pt}
    \end{tabular}
    \end{threeparttable}
    \end{adjustbox}
\caption{\textbf{Impact of threshold $\varphi$ used in positivity learning.} This experiment uses the DETR that detects $N$=100 objects in each frame. We explore $\varphi$ around $1/N$ = 0.010, \textit{i.e.}, the same attention to all objects.}
\label{tab:topk}
\end{table}

\begin{figure*}[!t]
\centering
 \includegraphics[width=0.85\textwidth]{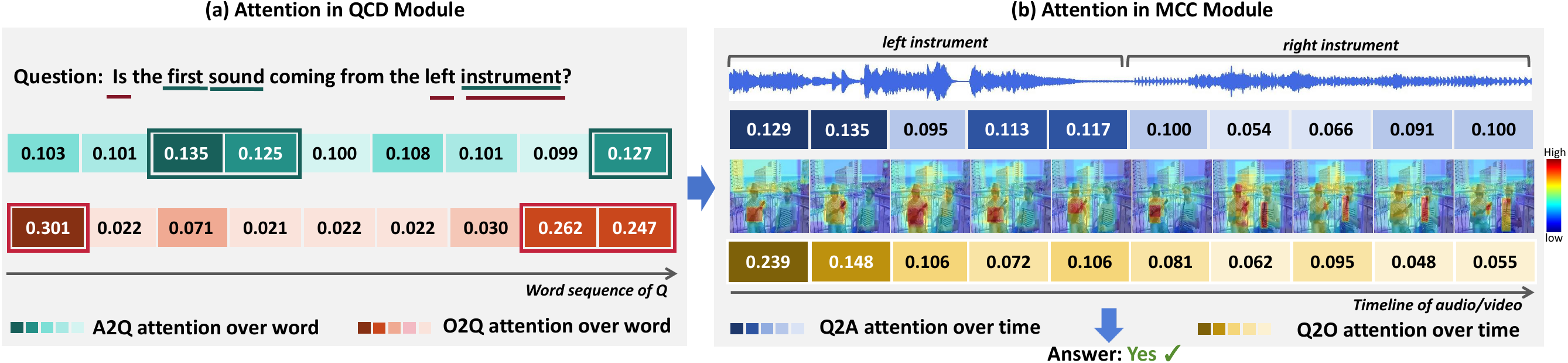}
  \caption{
   \textbf{Multi-modal attention visualization during answer reasoning process.} A2Q/O2Q from the QCD module is the abbreviation of audio/object to question, and Q2A/Q2O from the MCC module is the abbreviation of question to audio/object.}
  \label{Fig:attention}
\end{figure*}

\begin{figure*}[!t]
\centering
 \includegraphics[width=0.85\textwidth]{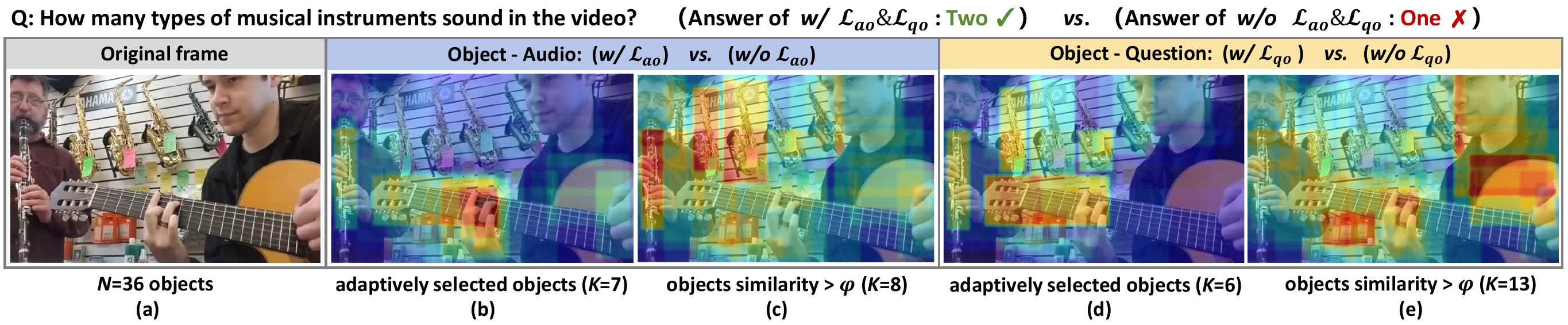}
  \caption{{{\textbf{The ablation visualization of adaptive positivity learning with $\mathcal{L}_{ao}$ and $\mathcal{L}_{qo}$.}}
  Two instruments make sounds in this example (a). With $\mathcal{L}_{ao}$ and $\mathcal{L}_{qo}$, these objects relevant to the audio and question are adaptively selected (b, d); Otherwise, the model highlights many irrelevant objects (c, e).}}
  \label{Fig:objects by losses}
\end{figure*}

\subsection{Qualitative Results}
In this subsection, we display some qualitative results to verify the effectiveness of the proposed APL method. The experiments are based on the Faster R-CNN object detector.

\noindent\textbf{Visualization of the Multi-modal Attention Weights.}
Our network uses the QCD and the MCC modules (Fig.~\ref{fig:framework}) for multi-modal interaction.
To explore whether these two modules actually learn the question-related audio-visual clues, we visualize the attention weights in the multi-head attention (Eq.~\ref{eq:TFM}) in these modules.
For the QCD module, we display the attention of each modality to each word of the question. As shown in Fig.~\ref{Fig:attention} (a), the audio modality has higher weights to these audio-related words in the question, \textit{e.g.}, ``\textit{first sound}''.
Also, the visual modality focuses more on those words relevant to the spatial location of visual objects, such as ``\textit{left instrument}''.
With QCD module, each modality successfully seizes the relevant keywords in the question.
Then in the subsequent MCC module, as shown in Fig.~\ref{Fig:attention} (b), the left instrument emits sound at the first five video segments, and the attention of the question to audio/visual successfully moves to these temporal segments.
Also, the heat maps in the video frames reflecting the question-object attention clue verify that the left instrument highly relevant to the question is highlighted in these earlier segments.
These results show that our model learns to understand what visual objects or sound clues the question cares about, and the model further collects highly relevant clues from each modality, ultimately predicting the correct answer ``\textit{Yes}''.

\noindent\textbf{Visualization of the Object-aware Adaptive Positivity Learning.}
To enhance the interpretability, 
we showcase an example to see what objects are adaptively selected with the objectives $\mathcal{L}_{qo}$ and $\mathcal{L}_{ao}$.
In Fig.~\ref{Fig:objects by losses} (a), we observe two men playing musical instruments (\textit{clarinet} and \textit{guitar}) in the scene, while some \textit{saxophones} hanging on the wall are silent.
In Fig.~\ref{Fig:objects by losses} (b)-(e), we visualize the bounding boxes of the selected objects based on the similarity of audio-object or question-object pairs.
Bright brown color indicates higher similarity values.
Fig.~\ref{Fig:objects by losses} (b) demonstrates the results with loss $\mathcal{L}_{ao}$, which adaptively selects the \textit{clarinet} and \textit{guitar} as positivity objects.
However, in the absence of $\mathcal{L}_{ao}$ constraint, the focus shifts to \textit{saxophone} hanging on the wall, as depicted in Fig.~\ref{Fig:objects by losses} (c).
This indicates that $\mathcal{L}_{ao}$ helps to effectively locate sound-related objects.
Similar phenomenons can be seen in Fig.~\ref{Fig:objects by losses} (d).
With loss $\mathcal{L}_{qo}$, the model successfully highlights the question-related objects, such as various musical instruments.
While without $\mathcal{L}_{qo}$, in Fig.~\ref{Fig:objects by losses} (e), the model pays large attention to more objects irrelevant to the question, such as the \textit{man} and \textit{wall decorations}.
With both $\mathcal{L}_{ao}$ and $\mathcal{L}_{qo}$, our model is able to identify the key objects not only related to the audio but also relevant to the question, thereby it can provide the correct answer ``\textit{Two}''.

\section{Conclusion}
In this paper, we propose to explore fine-grained object clues for the challenging audio-visual question answering task.
We present a multi-modal interaction network to encode the relations among the visual objects, audio, and question.
The core processing units are the question-conditioned clue discovery module and the modality-conditioned clue collection module.
The former module makes the audio/visual modality aware of the relevant keywords mentioned in the question, and the latter further highlights the informative clues contained in the audio/visual modality.
In addition, we also propose an object-aware adaptive-positivity learning strategy. 
It uses two contrastive loss functions to encourage the model to adaptively recognize the highly relevant question-object pairs and audio-object pairs.
Quantitative and qualitative experimental results verify that our method is able to seize the key clues in the question, audio, and visual objects, making it superior for audio-visual question answering.
\section*{Acknowledgements }
We would like to express our sincere gratitude to Dr. Liang Zheng for his invaluable comments and insightful suggestions, which helped us to improve this paper significantly.
This work was supported by the National Key R\&D Program of China (2022YFB4500600), the National Natural Science Foundation of China (62272144, 72188101, 61725203, 62020106007, and U20A20183), and the Major Project of Anhui Province (202203a05020011).

\bibliography{aaai24}
\end{document}